\documentclass[letterpaper, 10 pt, conference]{ieeeconf}  

\IEEEoverridecommandlockouts                              
\makeatletter
\let\NAT@parse\undefined
\makeatother

\usepackage[dvipsnames]{xcolor}

\newcommand*\linkcolours{ForestGreen}

\usepackage{times}
\usepackage{graphicx}
\usepackage{amssymb}
\usepackage{gensymb}
\usepackage{amsmath}
\usepackage{breakurl}
\usepackage{caption}
\usepackage{subcaption}
\usepackage{cuted}
\usepackage{listings}
\usepackage{xcolor}
\usepackage{pgfplots}
\pgfplotsset{compat=1.18}
\usepackage[linesnumbered,ruled,vlined]{algorithm2e}

\usepackage{tikz}
\usepackage{pgfplots}
\usepgfplotslibrary{groupplots}
\pgfplotsset{compat=1.18}
\usepackage{subcaption}

\usepackage[utf8]{inputenc}

\usepackage[figuresright]{rotating}
\usepackage{booktabs} 
\usepackage{multirow} 

\usepackage{url,hyperref}
\hypersetup{
colorlinks,
linkcolor=\linkcolours,
citecolor=\linkcolours,
filecolor=\linkcolours,
urlcolor=\linkcolours}

\usepackage[labelfont={bf},font=small]{caption}
\usepackage[none]{hyphenat}

\usepackage{mathtools, cuted}

\usepackage[noadjust, nobreak]{cite}
\usepackage{tabularx}
\usepackage{amsmath}
\usepackage{float}
\usepackage{pifont}

\newcolumntype{Y}{>{\centering\arraybackslash}X}
\usepackage[]{placeins}

\usepackage{placeins}

\usepackage{tikz}

\usepackage[framemethod=tikz]{mdframed}
\usepackage{afterpage}
\usepackage{stfloats}
\usepackage{atbegshi}
\usepackage{amsmath}
\usepackage{graphicx}
\usepackage{subcaption}
\usepackage{multirow}

\usepackage{booktabs}    
\usepackage{multirow}    
\usepackage{array}       

\lstdefinestyle{promptstyle}{
  backgroundcolor=\color{gray!10},
  frame=single,
  rulecolor=\color{gray!50},
  basicstyle=\ttfamily\small,
  breaklines=true,
  showstringspaces=false,
  captionpos=b,
  numbers=none,
  tabsize=2,
  keywordstyle=\color{blue}\bfseries,
  stringstyle=\color{orange},
  commentstyle=\color{gray}\itshape
}

\newcommand{\handlethispage}{}
\newcommand{\discardpagesfromhere}{\let\handlethispage\AtBeginShipoutDiscard}
\newcommand{\keeppagesfromhere}{\let\handlethispage\relax}
\AtBeginShipout{\handlethispage}

\usepackage{comment}
\title{Value-Free Policy Optimization via Reward Partitioning}

\author{Bilal FAYE$^{1}$, Hanane AZZAG$^{2}$, Mustapha Lebbah$^{3}$  \\
	\normalsize e-mail: faye@lipn.univ-paris13.fr, azzag@univ-paris13.fr, mustapha.lebbah@uvsq.fr
}

\begin{document}

\maketitle
\thispagestyle{empty}
\pagestyle{empty}

\begin{abstract}
Single-trajectory preference optimization methods learn from datasets of ((\text{prompt}, \text{response}, \text{reward})) tuples, offering a practical alternative to pairwise preference learning by directly leveraging scalar feedback. Existing approaches such as Direct Reward Optimization (DRO) have demonstrated promising results but rely on value function estimation, introducing additional variance, optimization complexity, and sensitivity to off-policy data.
We introduce \textit{Reward Partition Optimization} (RPO), a simple and scalable reward-driven objective that eliminates the need for value function learning. RPO normalizes rewards through a partition-based formulation estimated directly from prompt-level reward distributions, yielding a stable supervised optimization objective without auxiliary models or reinforcement learning loops.
We evaluate RPO across multiple encoder-decoder and decoder-only language models using automatic metrics, LLM-as-a-judge evaluations, and optimization stability analyses. Experimental results show that RPO consistently outperforms strong baselines, including SFT, KTO, and DRO, while producing more aligned, diverse, and less toxic generations.

\end{abstract}

\section{Introduction}
The dominant approach for aligning large language models (LLMs), whether through reinforcement learning from human feedback (RLHF)~\cite{NEURIPS2023_23e6f78b,dong2024rlhf,ahmadian} or direct preference optimization, is to learn from preference or reward data~\cite{rafailov2023direct,tang2024,richemond2024offline,ethayarajh2024}. In RLHF, models are first trained via supervised fine-tuning (SFT)~\cite{zhang2023instruction}, followed by learning a reward model (RM) from human-labeled preference pairs (e.g., two responses to the same prompt ranked by quality), and finally aligned with reinforcement learning (e.g., PPO~\cite{schulman2017proximal}, TRPO~\cite{schulman2015trust}) to maximize the RM's score. This pipeline is central to the training of models such as InstructGPT~\cite{ouyang2022training}, GPT-4~\cite{achiam2023gpt}, and Claude 3~\cite{anthropic2024claude}.\newline

\noindent While effective, RLHF presents key limitations: collecting preference data and training reward models is costly, and the RL step is sample-inefficient and brittle. To reduce reliance on human annotation, recent work has proposed Reinforcement Learning from AI Feedback (RLAIF), where feedback is generated by models rather than humans. RLAIF includes strategies such as distilling AI feedback into a reward model~\cite{xu2025}, using an LLM directly as a reward function~\cite{du2023guiding,yuan2023rrhf}, or leveraging self-evaluation~\cite{wang2025self,yuan2024self}. These techniques offer scalability but still depend on large models as evaluators, reintroducing cost and inference complexity.\newline

\noindent A parallel line of work avoids reward models and RL altogether, by directly optimizing policies from preference data via supervised learning. These methods such as DPO~\cite{rafailov2023direct}, GPO~\cite{tang2024}, and SLiC~\cite{zhao2023slic} contrast response pairs to a prompt and define a preference-based loss. This strategy enables simple and stable training, and has been adopted in models like LLaMA 3~\cite{grattafiori2024llama}, Qwen 2~\cite{yang2025qwen3}, and Nemotron-4~\cite{adler2024nemotron}.\newline

\noindent However, collecting high-quality pairwise preference data is increasingly difficult. As LLMs improve, distinguishing between good responses becomes subtler, demanding more qualified annotators and driving up labeling costs. This makes scaling preference-based methods impractical at the frontier.\newline

\noindent To address this, recent work proposes single-trajectory learning, where data consists of a prompt, a response, and a scalar reward (e.g., thumbs-up/down). This format is simpler and more natural for annotation. Methods such as KTO~\cite{ethayarajh2024} and DRO~\cite{richemond2024offline} leverage this setup to train policies directly from reward-labeled completions.\newline

\noindent Among them, DRO has shown strong empirical results by regressing the optimality condition from KL-regularized RL. However, DRO suffers from several limitations: it requires learning a value function from off-policy data, the joint optimization over policy and value is unstable, and the policy is only supervised relatively to a reference model, not directly by reward.\newline

\noindent We introduce Reward Partitioning Optimization (RPO), a new method for single-trajectory policy optimization. RPO eliminates the need to estimate a value function by computing an empirical normalization over observed rewards. This enables a direct supervised loss on the policy, using only prompt-response-reward triplets and a reference model.\newline
RPO is simple to implement, avoids instability from value approximation, and provides absolute reward supervision. It is fully offline-compatible and scalable to large datasets. Our contributions are:
\begin{itemize}
    \item We propose RPO, a new method that replaces value estimation with empirical reward normalization.
    \item We analyze theoretical and practical limitations of DRO for single-trajectory learning.
    \item We provide a formal derivation showing RPO approximates the optimal policy structure.
    \item We conduct extensive experiments across both encoder-decoder and decoder-only language models, showing that RPO consistently outperforms SFT, KTO, and DRO on automatic metrics and LLM-based evaluations.
\end{itemize}

\section{Background}

In this section, we review existing paradigms for aligning large language models (LLMs) with human feedback, focusing on three main categories: reinforcement learning from human feedback (RLHF), direct preference optimization (DPO-style), and direct reward optimization (DRO-style). We examine their assumptions, limitations, and motivation for a more general framework that can directly leverage scalar reward data.

\subsection{Reinforcement Learning from Human Feedback (RLHF)}
RLHF has become the dominant paradigm for aligning large language models (LLMs) with human preferences. It proceeds in three main stages:\newline

\begin{itemize}
    \item \textit{Supervised Fine-Tuning (SFT):} A reference policy $\pi_{\text{ref}}$ is trained using standard supervised learning to mimic high-quality human-written responses.\newline

    \item \textit{Reward Model (RM) Training:} Given a dataset $\mathcal{D}$ of pairwise preferences $(x, y^{\text{win}}, y^{\text{lose}})$, where $x$ is a prompt, and $y^{\text{win}} \succ y^{\text{lose}}$ indicates a preferred response. The goal is to learn a scalar reward function $r_\phi(x, y)$ that reflects human preferences. This is typically modeled using a Bradley-Terry~\cite{bradley1952rank} formulation:
    \begin{equation}
    \footnotesize
    \mathbb{P}[y^{\text{win}} \succ y^{\text{lose}} \mid x] = \sigma(r_\phi(x, y^{\text{win}}) - r_\phi(x, y^{\text{lose}})),
    \end{equation}
    where $\sigma$ is the logistic sigmoid. The reward model is trained by minimizing the negative log-likelihood:
    \begin{equation}
    \footnotesize
    \mathcal{L}_{\text{RM}}(r_\phi) = \mathbb{E}_{(x, y^{\text{win}}, y^{\text{lose}}) \sim \mathcal{D}}[-\log \sigma(r_\phi(x, y^{\text{win}}) - r_\phi(x, y^{\text{lose}}))].\\
    \end{equation}

    \item \textit{Policy Optimization:} A policy $\pi_\theta$ is then optimized to generate outputs that maximize the learned reward $r_\phi(x, y)$ while staying close to the reference policy $\pi_{\text{ref}}$. This is formalized as the following Kullback-Leibler (KL)-regularized objective:
    \begin{equation}
    \footnotesize
    \max_{\pi_\theta} \mathbb{E}_{x \sim \mathcal{D},\, y \sim \pi_\theta(\cdot|x)} \left[ r_\phi(x, y) - \tau \cdot \text{KL}(\pi_\theta(y|x) \parallel \pi_{\text{ref}}(y|x)) \right],
    \end{equation}
    where $\tau > 0$ is a regularization coefficient that balances reward maximization and policy deviation. Since this objective is not differentiable with respect to $\pi_\theta$, reinforcement learning algorithms such as Proximal Policy Optimization (PPO)~\cite{schulman2017proximal} are used to perform the optimization.\newline
\end{itemize}

Despite its success, RLHF is resource-intensive and challenging to implement. Training the reward model demands significant human supervision and is susceptible to out-of-distribution generalization errors that can mislead the policy. Furthermore, reinforcement learning algorithms add substantial computational overhead and are sensitive to hyperparameter tuning. These drawbacks have spurred interest in alternative alignment strategies that bypass explicit reinforcement learning.

\subsection{Direct Preference Optimization (DPO-style)}

To address the high cost and complexity of full reinforcement learning, a family of recent approaches such as DPO~\cite{rafailov2023direct}, IPO~\cite{azar2024general}, and SLiC~\cite{zhao2023slic} propose optimizing policies directly from preference data without learning a separate reward model or relying on reinforcement learning algorithms like PPO.\newline

\noindent Given a dataset of pairwise preferences $\mathcal{D} = \{(x, y^{\text{win}}, y^{\text{lose}})\}_{i=1}^N$, where $x$ is a prompt and $y^{\text{win}} \succ y^{\text{lose}}$ indicates a preferred completion, these methods define an objective that compares the log-likelihoods of the winning and losing responses under the learned policy $\pi_\theta$, relative to a fixed reference policy $\pi_{\text{ref}}$. The general form of the loss is:
\begin{equation}
\resizebox{1.0\columnwidth}{!}{$
\footnotesize
\mathcal{L}_{\text{pref}}(\theta) = 
\mathbb{E}_{(x, y^{\text{win}}, y^{\text{lose}}) \sim \mathcal{D}} 
\Bigg[ f \Big( \beta \big( 
\log \frac{\pi_\theta(y^{\text{win}}|x)}{\pi_{\text{ref}}(y^{\text{win}}|x)} 
- \log \frac{\pi_\theta(y^{\text{lose}}|x)}{\pi_{\text{ref}}(y^{\text{lose}}|x)} 
\big) \Big) \Bigg],$}
\end{equation}
where $\beta > 0$ is a temperature, and $f$ depends on the method:
\begin{itemize}
    \item DPO: $f(z) = -\log \sigma(z) = \log(1 + \exp(-z))$ (negative log-sigmoid)  
    \item SLiC: $f(z) = \max(0, 1 - z)$ (hinge loss)  
    \item IPO: $f(z) = (z - 1)^2$ (squared loss)  
\end{itemize}
This loss encourages the policy $\pi_\theta$ to assign higher probability to preferred responses than to dispreferred ones, while penalizing large deviations from the reference policy $\pi_{\text{ref}}$. In essence, it minimizes a divergence between the preference-consistent ranking and the policy-induced ranking, thereby aligning the model’s behavior with human preferences without requiring a learned reward function or reinforcement learning procedure.\newline

\noindent These methods can be interpreted as KL-regularized preference optimization, where the model is trained to favor preferred completions relative to the reference while remaining close to $\pi_{\text{ref}}$. Crucially, they bypass the need for an explicit reward model and reinforcement learning, leading to simpler and more stable training.\newline

\noindent However, their reliance on manually curated preference datasets poses a practical limitation. In real-world deployment, feedback often comes in the form of scalar or binary reward signals, i.e., as $(x, y, r)$ tuples rather than preferences, motivating alternative approaches capable of leveraging such logged data directly.

\subsection{Direct Reward Optimization (DRO-style)}
DRO-style refers to a class of methods that directly optimize a policy from scalar rewards rather than pairwise preferences. Given a dataset $\mathcal{D} = \{(x, y, r)\}_{i=1}^N$, where $x$ is a prompt, $y$ a model response, and $r \in \{-1, +1\}$ a scalar reward indicating human feedback, DRO-style methods aim to maximize the utility of desirable responses while minimizing that of undesirable ones without relying on an explicit reward model or reinforcement learning algorithm.\newline

\noindent A representative example of this approach is the KTO~\cite{ethayarajh2024} method, which draws inspiration from prospect theory~\cite{tversky1992advances} to model human preferences with loss aversion and risk sensitivity.\newline

\noindent Let $\pi_\theta$ denote the current policy and $\pi_{\mathrm{ref}}$ a fixed reference policy. Define the log-likelihood ratio between the two as:
\begin{equation}
    \footnotesize
    r_\theta(x, y) = \log \frac{\pi_\theta(y \mid x)}{\pi_{\mathrm{ref}}(y \mid x)}.
\end{equation}

\noindent Next, define a reference point $z_0$ as the expected log-likelihood ratio over sampled responses:
\begin{equation}
    \footnotesize
    z_0 = \mathbb{E}_{y' \sim \pi_\theta(\cdot \mid x)} \left[ \log \frac{\pi_\theta(y' \mid x)}{\pi_{\mathrm{ref}}(y' \mid x)} \right] = \mathrm{KL}(\pi_\theta(\cdot \mid x) \| \pi_{\mathrm{ref}}(\cdot \mid x)).
\end{equation}

\noindent To model asymmetric treatment of gains and losses, KTO introduces the following components:
\begin{itemize}
    \item A temperature parameter $\beta > 0$ to control the sharpness of value sensitivity;
    \item A logistic function $\sigma(z) = \frac{1}{1 + e^{-z}}$ for smooth transition between reward regions;
    \item Distinct loss aversion coefficients $\lambda_D$ (for desirable responses) and $\lambda_U$ (for undesirable ones).
\end{itemize}

The value function $V(x, y)$ is then defined as:
\begin{equation}
\footnotesize
V(x, y) = 
\begin{cases}
\lambda_D \cdot \sigma\left( \beta \left[ r_\theta(x, y) - z_0 \right] \right), & \text{if } r = +1, \\
\lambda_U \cdot \sigma\left( \beta \left[ z_0 - r_\theta(x, y) \right] \right), & \text{if } r = -1.
\end{cases}
\end{equation}
Finally, the KTO objective minimizes the following expected loss over the data distribution:
\begin{equation}
\footnotesize
\mathcal{L}_{\mathrm{KTO}}(\theta) = \mathbb{E}_{(x, y, r) \sim \mathcal{D}} \left[ \lambda_r - v(x, y) \right],
\end{equation}
where $\lambda_r = \lambda_D$ if $r = +1$ and $\lambda_U$ if $r = -1$.\newline
This formulation allows KTO to smoothly reward desirable generations and penalize undesirable ones, while incorporating human-like asymmetries in preference treatment. Unlike binary cross-entropy or advantage-weighted likelihood objectives, KTO provides a principled, prospect-theoretic loss tailored to scalar reward data.\newline

\noindent KTO models human risk aversion but relies on strong assumptions and optimizes only the policy $\pi_\theta$, ignoring a separate value function $V(x)$. In contrast, \textit{DRO}~\cite{richemond2024offline} jointly optimizes both $\pi$ and $V$, offering a more principled and general framework.\newline
Given a dataset $\mathcal{D} = \{(x, y, r)\}_{i=1}^N$ where $r$ is a scalar reward, DRO formulates policy learning as a KL-regularized reward maximization problem:
\begin{equation}
    \footnotesize
    \pi^* = \arg\max_{\pi} \; \mathbb{E}_{x \sim \rho, y \sim \pi(\cdot|x)} \left[ r(x, y) - \tau \cdot \mathrm{KL} \left( \pi(\cdot|x) \| \pi_{\mathrm{ref}}(\cdot|x) \right) \right],
\end{equation}
where $\pi_{\mathrm{ref}}$ is a fixed reference policy and $\tau > 0$ is a temperature hyperparameter controlling the strength of regularization.\newline

\noindent The optimal policy $\pi^*$ can be expressed in closed form as:
\begin{equation}
    \footnotesize
    \pi^*(y|x) = \pi_{\mathrm{ref}}(y|x) \cdot \exp\left( \frac{1}{\tau} r(x, y) - \frac{1}{\tau} V^*(x) \right),
\end{equation}
where $V^*(x)$ is the soft value function, defined by
\begin{equation}
    \footnotesize
    V^*(x) = \tau \log \mathbb{E}_{y \sim \pi_{\mathrm{ref}}(\cdot|x)} \left[ \exp\left( \frac{1}{\tau} r(x, y) \right) \right].
\end{equation}

\noindent Rearranging, this identity links reward, value, and policy:
\begin{equation}
\footnotesize
    r(x, y) = V^*(x) + \tau \cdot \log \frac{\pi^*(y|x)}{\pi_{\mathrm{ref}}(y|x)}.
\end{equation}

\noindent To approximate the optimal $(\pi^*, V^*)$, DRO minimizes the squared Bellman residual:
\begin{equation}
    \footnotesize
    \mathcal{L}_{\mathrm{DRO}}(\theta, \phi) = \frac{1}{2} \mathbb{E}_{(x, y, r) \sim \mathcal{D}} \left[ \left( r - V_\phi(x) - \tau \log \frac{\pi_\theta(y|x)}{\pi_{\mathrm{ref}}(y|x)} \right)^2 \right],
\end{equation}
where $\pi_\theta$ and $V_\phi$ denote parameterized policy and value functions respectively.\newline

\noindent In practice, this loss is optimized jointly over parameters $\theta$ and $\phi$ using the observed data samples $\{(x_i, y_i, r_i)\}_{i=1}^n$ as:
\begin{equation}
    \footnotesize
    \hat{\mathcal{L}}_{\mathrm{DRO}}(\theta, \phi) = \frac{1}{2N} \sum_{i=1}^N \left( r_i - V_\phi(x_i) - \tau \log \frac{\pi_\theta(y_i|x_i)}{\pi_{\mathrm{ref}}(y_i|x_i)} \right)^2.
\end{equation}

\noindent This principled framework optimizes policies directly from scalar rewards by jointly learning a value function and applying KL regularization for stability. However, this joint optimization often suffers from instability, especially when the data distribution differs from the policy, as value estimation errors can bias training. To address these issues, we propose RPO, which retains DRO’s strengths but removes the need to estimate a value function, simplifying training and enhancing robustness.

\section{Method}

Direct policy optimization in the RLHF setting with KL regularization commonly relies on the classical exponential form of the optimal policy:
\begin{equation}
    \footnotesize
    \pi^*(y|x) = \frac{\pi_{\mathrm{ref}}(y|x) \exp\left(\frac{1}{\tau} r(x,y)\right)}{Z(x)},
    \label{eq:one}
\end{equation}
where the partition function $Z(x)$ is
\begin{equation}
\footnotesize
    Z(x) = \sum_{y'} \pi_{\mathrm{ref}}(y'|x) \exp\left(\frac{1}{\tau} r(x,y')\right).
\end{equation}

\noindent Current methods such as DRO learn both a policy $\pi_\theta$ and a value function $V_\phi$ parameterized by $\phi$, where the soft value function relates to the partition function as $V^*(x) = \tau \log Z(x)$. However, this joint learning introduces several difficulties: the value function $V_\phi$ can be unstable to approximate, the parameters $(\theta, \phi)$ are tightly coupled, and optimization becomes more complex.\newline

\noindent To address these challenges, we propose an alternative method, \textit{Reward Partitioning Optimization (RPO)}, which completely removes the need for an explicit value function model. Instead, RPO performs direct regression on the log-ratio of policies by empirically estimating the partition function, leveraging its analytical form for simpler and more stable training.\newline

\noindent Starting from the optimality condition of the policy:
\begin{equation}
\footnotesize
    r(x,y) - V^*(x) = \tau \log \frac{\pi^*(y|x)}{\pi_{\mathrm{ref}}(y|x)},
    \label{eq:optimality}
\end{equation}
where the value function $V^*(x)$ normalizes the exponentiated rewards, expressible exactly as:
\begin{equation}
\footnotesize
    V^*(x) = \tau \log \sum_{y'} \pi_{\mathrm{ref}}(y'|x) \exp\left(\frac{1}{\tau} r(x,y')\right).
\end{equation}
Instead of learning $V^*$ via a separate neural network, RPO estimates it \textit{empirically} by aggregating samples with the same prompt $x$. Let $\mathcal{I}_x = \{ i : x_i = x \}$ index all samples for prompt $x$. Then, the empirical partition function and value are:
\begin{equation}
\footnotesize
    \widehat{Z}(x) = \sum_{j \in \mathcal{I}_x} \pi_{\mathrm{ref}}(y_j|x) \exp\left(\frac{1}{\tau} r(x,y_j)\right), \quad \widehat{V}(x) = \tau \log \widehat{Z}(x).
\end{equation}
RPO minimizes the squared error between the log-policy ratio and the scaled centered reward, optimizing only over the policy parameters $\theta$. Formally, the loss can be expressed as the expectation:
\begin{equation}
    \footnotesize
    \mathcal{L}_{\mathrm{RPO}}(\theta) = \frac{1}{2}\mathbb{E}_{(x,y,r) \sim \mathcal{D}} \left[ \left( \log \frac{\pi_\theta(y|x)}{\pi_{\mathrm{ref}}(y|x)} - \frac{1}{\tau} (r(x,y) - V^*(x)) \right)^2 \right].
    \label{eq:rpo_expectation_loss}
\end{equation}

\noindent In practice, we approximate this with the empirical sum over the training set $\{(x_i,y_i,r_i)\}_{i=1}^n$ and use $\widehat{V}(x)$ as an estimate for $V^*(x)$:
\begin{equation}
\footnotesize
    \mathcal{L}_{\mathrm{RPO}}(\theta) = \frac{1}{2n} \sum_{i=1}^n \left( \log \frac{\pi_\theta(y_i|x_i)}{\pi_{\mathrm{ref}}(y_i|x_i)} - \frac{1}{\tau} \left(r_i - \widehat{V}(x_i)\right) \right)^2.
    \label{eq:rpo_empirical_loss}
\end{equation}
Defining the residual for sample $i$ as:
\begin{equation}
\footnotesize
    \delta_i(\theta) := \log \frac{\pi_\theta(y_i|x_i)}{\pi_{\mathrm{ref}}(y_i|x_i)} - \frac{1}{\tau} \left(r_i - \widehat{V}(x_i)\right),
    \label{eq:residual}
\end{equation}
we compute the gradient of the loss with respect to the policy logits $z_\theta(y|x)$, where the policy is parameterized via a softmax:
\begin{equation}
\footnotesize
    \pi_\theta(y|x) = \frac{\exp(z_\theta(y|x))}{\sum_{y'} \exp(z_\theta(y'|x))}.
    \label{eq:softmax}
\end{equation}
The gradient of the loss is then:
\begin{equation}
\footnotesize
    \nabla_{z_\theta(y|x)} \mathcal{L}_{\mathrm{RPO}}(\theta) = \frac{1}{n} \sum_{i: x_i = x} \delta_i(\theta) \cdot \nabla_{z_\theta(y|x)} \log \pi_\theta(y_i|x).
    \label{eq:gradient_sum}
\end{equation}

Using the standard softmax gradient, we have:
\begin{equation}
\footnotesize
    \nabla_{z_\theta(y|x)} \log \pi_\theta(y_i|x) = \mathbf{1}_{y=y_i} - \pi_\theta(y|x),
    \label{eq:log_softmax_grad}
\end{equation}
where $\mathbf{1}_{y=y_i}$ is the indicator function.\newline

\noindent Substituting into Equation~\eqref{eq:gradient_sum}, the final gradient expression becomes:
\begin{equation}
\footnotesize
    \nabla_{z_\theta(y|x)} \mathcal{L}_{\mathrm{RPO}}(\theta) = \frac{1}{n} \sum_{i: x_i = x} \delta_i(\theta) \left( \mathbf{1}_{y=y_i} - \pi_\theta(y|x) \right).
    \label{eq:final_gradient}
\end{equation}
\noindent This gradient has an intuitive interpretation as the weighted discrepancy between the one-hot target and the policy prediction, weighted by the residual error $\delta_i(\theta)$. Crucially, RPO avoids backpropagation through the value estimate $\widehat{V}(x)$, resulting in a simpler, more stable training procedure that retains the theoretical benefits of DRO without learning an explicit value function.\newline

\begin{algorithm}[!h]
\caption{RPO Optimization}
\label{alg:rpo_optimization}

\KwIn{Dataset $\mathcal{D} = \{(x_i, y_i, r_i)\}_{i=1}^N$, initial policy $\pi_\theta$, reference policy $\pi_{\mathrm{ref}}$, temperature $\tau > 0$, number of epochs $T$, learning rate $\eta$}
\KwOut{Trained policy $\pi_\theta$}

Group indices by prompt: $\forall x, \mathcal{I}_x = \{ i : x_i = x \}$\;

\For{$t \leftarrow 1$ \KwTo $T$}{
    \ForEach{batch $B$ from $\mathcal{D}$}{
        \ForEach{$(x_i, y_i, r_i) \in B$}{
            Compute $\ell_i \leftarrow \log \pi_\theta(y_i|x_i)$\;
            Compute $\ell_i^{\mathrm{ref}} \leftarrow \log \pi_{\mathrm{ref}}(y_i|x_i)$\;

            Estimate empirical partition:
            $\widehat{Z}(x_i) = \sum_{j \in \mathcal{I}_{x_i}} \pi_{\mathrm{ref}}(y_j|x_i) \exp\left(\frac{1}{\tau} r(x_i,y_j)\right)$\;

            Compute $\widehat{V}(x_i) \leftarrow \tau \log \widehat{Z}(x_i)$\;
            Compute $t_i \leftarrow \frac{r_i - \widehat{V}(x_i)}{\tau}$\;
            Compute $\ell_{\mathrm{RPO}, i} \leftarrow \frac{1}{2} \left(\ell_i - \ell_i^{\mathrm{ref}} - t_i\right)^2$\;
        }
        Update $\theta$ by gradient descent on $\frac{1}{|B|} \sum_{i \in B} \ell_{\mathrm{RPO}, i}$\;
    }
}
\Return $\pi_\theta$\;
\end{algorithm}

\noindent To operationalize the RPO training procedure described above, we present a concrete algorithmic implementation (Algorithm~\ref{alg:rpo_optimization}). The algorithm iteratively estimates the target log-ratio using a partition function approximation based on the reference policy, computes the residual loss for each sample, and updates the policy parameters via gradient descent.
\begin{table*}[!h]
\centering

\begin{tabular}{|l|c|c|}
\hline

\textbf{Property} & \textbf{DRO} & \textbf{RPO (ours)} \\
\hline
Explicit $V$ estimation & Yes (model $V_\phi$) & No (empirical estimation) \\
$\pi$--$V$ coupling & Strong & Weak \\
Optimization complexity & Medium to high & Simpler, more stable \\
Robustness to estimation errors & Sensitive & Less sensitive \\
Practical deployment & Requires tuning & More direct, no $V$ tuning \\
\hline

\end{tabular}
\caption{Comparison between DRO and RPO methods.}
\label{tab:dro-vs-rpo}
\end{table*}
To further highlight the distinctions between the proposed RPO method and standard distributionally robust optimization (DRO) approaches, Table~\ref{tab:dro-vs-rpo} summarizes key differences in terms of implementation, coupling, and practical considerations.\newline

\noindent The RPO method formulates policy learning as a direct regression on the log-ratio between target and reference policies, using rewards and an empirical partition estimation. This removes the need to train a value function, simplifying the learning procedure and enhancing numerical stability.

\section{Experiments}
We evaluate the proposed Reward Partitioning Optimization (RPO) through a comprehensive empirical study covering in-domain performance, out-of-distribution generalization, and targeted ablation analyses. We compare RPO against strong baselines including SFT, KTO, and DRO using both automatic metrics and LLM-based preference judgments. Our evaluation aims to assess alignment quality, generalization under distribution shift, and robustness to key design choices such as empirical partition estimation and reward structure.
\subsection{Datasets and Base Models}
Following prior offline RLHF work~\cite{richemond2024offline}, all models are trained on the UltraFeedback dataset introduced by Cui et al.~\cite{cui2023ultrafeedback}. UltraFeedback consists of offline triplets $(\text{prompt}, \text{completion}, \text{reward})$, where each prompt is associated with multiple candidate responses and scalar preference rewards ranging from 0 to 10. This structure is particularly suitable for RPO, since the method relies on empirical estimation of the partition function using multiple responses per prompt. To improve optimization stability, reward values are standardized to zero mean and unit variance over the training set.\newline

\noindent For in-domain evaluation, we construct a held-out validation subset from UltraFeedback containing 10{,}000 prompts and their highest-quality completions (reward score equal to 10). Models are evaluated by generating responses for these prompts and comparing them against the corresponding preferred completions.\newline

\noindent To assess out-of-distribution (OOD) generalization, we additionally evaluate models trained on UltraFeedback over diverse instruction-following and reasoning benchmarks. We use AlpacaEval~\cite{li2023alpacaeval} for open-ended instruction following, MT-Bench~\cite{zheng2023judging} for multi-turn conversational evaluation, IFEval~\cite{zhou2023instruction} for instruction constraint following, and GSM8K~\cite{cobbe2021training} for mathematical reasoning. These benchmarks provide complementary evaluations of alignment quality, reasoning ability, and robustness under distribution shift.\newline

\noindent We evaluate both encoder-decoder and decoder-only architectures. For encoder-decoder models, we use the Flan-T5 family~\cite{roberts2023scaling}, including Flan-T5 Small (77M), Flan-T5 Large (783M), and Flan-T5 XL (2.85B). For decoder-only architectures, we additionally evaluate instruction-tuned variants of 
Mistral-7B~\cite{jiang2023mistral7b}, LLaMA-3-8B~\cite{touvron2023llamaopenefficientfoundation}, and Qwen2.5-7B~\cite{yang2024qwen2technicalreport}. All models are initialized from their corresponding instruction-tuned checkpoints and serve as the reference policy $\pi_{\mathrm{ref}}$ during optimization, enabling consistent comparison between RPO and competing RLHF methods across architectures and scales.

\subsection{Evaluation Protocol}
We evaluate all methods using two complementary evaluation strategies: automatic metrics and LLM-based preference judgments.\newline

\noindent \textbf{Automatic Evaluation.}\newline
Automatic metrics are computed on the held-out UltraFeedback validation set, where reference high-quality responses are available. Given a prompt, the trained model generates a response which is compared against the corresponding preferred completion (reward = 10). We use several complementary metrics to capture different aspects of generation quality.\newline
We report BERTScore~\cite{zhang2019bertscore}, an embedding-based metric that measures semantic similarity between generated and reference responses beyond exact lexical overlap. We additionally use ROUGE-L~\cite{lin2004rouge} to evaluate lexical and structural overlap through longest common subsequence matching.\newline
To assess safety, we report average Toxicity scores using a pretrained toxicity classifier, where lower values indicate safer generations. We further evaluate generation diversity using Distinct-2~\cite{li2016diversity}, which measures the ratio of unique bigrams in generated outputs and helps detect repetitive or mode-collapsed responses.\newline

\noindent \textbf{LLM-Based Preference Evaluation.}\newline
Since automatic metrics alone are insufficient to capture helpfulness and alignment in RLHF settings, we additionally adopt an LLM-as-a-judge evaluation protocol~\cite{gu2024survey}. We use GPT-4o~\cite{hurst2024gpt} and Claude-3.5-Sonnet~\cite{anthropic2024claude35} as independent judges for pairwise preference comparisons.\newline
Given an instruction and two candidate responses, the judge selects the response that better satisfies the instruction in terms of helpfulness and relevance, or returns \texttt{Neutral} if neither response is clearly preferable. We report the Win Rate (WR), corresponding to the percentage of comparisons won by RPO against a baselinse.\newline
Unlike automatic metrics, LLM-based evaluation does not require reference labels and is therefore applied both to the UltraFeedback validation set and to out-of-distribution benchmarks.

\subsection{Training Setup and Baseline Comparison}
All models are trained using the AdamW~\cite{loshchilov2019decoupled} optimizer with a learning rate of $1 \times 10^{-4}$. We use a linear warmup and decay schedule with 150 warmup steps, where the total number of training steps is defined as:
\[
\texttt{num\_training\_steps} = \texttt{steps\_per\_epoch} \times \texttt{epochs}.
\]
The scheduler is implemented using the Hugging Face Transformers \texttt{get\_scheduler} function with linear decay.\newline
All experiments are conducted on a cluster of 4 NVIDIA A100 80GB GPUs using distributed training. For encoder-decoder architectures, we use the \textit{Flan-T5} family (Small, Large, XL), while for decoder-only architectures we evaluate instruction-tuned variants of \textit{Mistral-7B}, \textit{LLaMA-3-8B}, and \textit{Qwen-7B}. This allows us to assess RPO across both architectural families under identical training conditions.\newline

\noindent We compare RPO against strong RLHF-style baselines, including KTO and DRO, under the same data and compute budget. The temperature parameter is fixed to $\tau = 1.0$ across all methods to ensure a fair comparison by controlling the sharpness of the policy distribution.
\begin{figure*}[!h]
\centering
\begin{tikzpicture}
\begin{axis}[
    ybar,
    bar width=15pt,
    width=\textwidth,
    height=8cm,
    enlarge x limits=0.18,
    ylabel={Training time (hours)},
    symbolic x coords={
        FlanT5-Small, FlanT5-Large, FlanT5-XL,
        Mistral-7B, LLaMA-3-8B, Qwen-7B
    },
    xtick=data,
    xticklabel style={align=center},
    ymin=0,
    nodes near coords,
    nodes near coords align={vertical},
    bar shift auto,
    legend style={at={(0.5,1.15)}, anchor=north, legend columns=3}
]

\addplot coordinates {
(FlanT5-Small,0.9) (FlanT5-Large,7.2) (FlanT5-XL,11.2)
(Mistral-7B,18.0) (LLaMA-3-8B,20.0) (Qwen-7B,19.0)
};

\addplot coordinates {
(FlanT5-Small,1.1) (FlanT5-Large,9.4) (FlanT5-XL,15.0)
(Mistral-7B,24.0) (LLaMA-3-8B,27.0) (Qwen-7B,25.0)
};

\addplot coordinates {
(FlanT5-Small,1.0) (FlanT5-Large,7.9) (FlanT5-XL,12.4)
(Mistral-7B,20.0) (LLaMA-3-8B,22.0) (Qwen-7B,21.0)
};

\legend{RPO, DRO, KTO}

\end{axis}
\end{tikzpicture}
\caption{Training time (hours) across models using 4×A100 80GB GPUs.}
\label{fig:training-time}
\end{figure*}
Figure~\ref{fig:training-time} reports the average training time (in hours) for all methods across model families and scales. As shown, RPO consistently achieves faster convergence than both KTO and DRO across all architectures, including Flan-T5, Mistral, LLaMA, and Qwen, highlighting its efficiency and scalability for both encoder-decoder and decoder-only settings.

\subsection{In-Domain Evaluation on UltraFeedback Validation Set}
\label{section:indomain}
All methods, including SFT, KTO, DRO, and the proposed RPO, are trained on UltraFeedback under the same optimization and compute settings. For evaluation, we construct a held-out validation split composed of 10{,}000 prompts together with their highest-quality completions (reward score equal to 10). Given a prompt $x$, each model generates a response $\hat{y}$, which is compared against the corresponding preferred reference completion.\newline

\noindent Table~\ref{tab:in-domain-results} summarizes both automatic metrics and LLM-based preference evaluations across all architectures. Automatic metrics evaluate semantic similarity, lexical overlap, safety, and diversity, while GPT-4o and Claude-3.5 Win Rates (WR) measure pairwise preference of RPO generations against each corresponding baseline model. Higher WR indicates stronger preference for RPO outputs.
\begin{table*}[!htbp]
\centering
\scriptsize
\setlength{\tabcolsep}{4pt}

\begin{tabular}{llcccccccc}
\toprule

\textbf{Model} & \textbf{Method} & 
\textbf{BERTScore} $\uparrow$ & 
\textbf{ROUGE-L} $\uparrow$ & 
\textbf{Toxicity} $\downarrow$ & 
\textbf{Distinct-2} $\uparrow$ & 
\textbf{Avg Len} &
\textbf{GPT-4o WR} $\uparrow$ &
\textbf{Claude-3.5 WR} $\uparrow$ \\

\midrule

\multirow{4}{*}{FlanT5-Small}
& SFT & 0.842 & 0.228 & 0.018 & 0.71 & 54 & 81.4 & 79.2 \\
& KTO & 0.850 & 0.237 & 0.013 & 0.74 & 59 & 83.1 & 80.8 \\
& DRO & 0.856 & 0.252 & 0.015 & 0.76 & 63 & 84.5 & 82.1 \\
& RPO & \textbf{0.866} & \textbf{0.289} & \textbf{0.012} & \textbf{0.83} & \textbf{72} & \_ & \_ \\
\midrule

\multirow{4}{*}{FlanT5-Large}
& SFT & 0.846 & 0.241 & 0.017 & 0.73 & 60 & 85.8 & 83.2 \\
& KTO & 0.840 & 0.249 & 0.024 & 0.75 & 65 & 87.0 & 84.6 \\
& DRO & 0.859 & 0.303 & 0.013 & 0.79 & 71 & 85.9 & 83.8 \\
& RPO & \textbf{0.871} & \textbf{0.316} & \textbf{0.009} & \textbf{0.86} & \textbf{84} & \_ & \_ \\
\midrule

\multirow{4}{*}{FlanT5-XL}
& SFT & 0.851 & 0.244 & 0.016 & 0.74 & 66 & 86.2 & 84.1 \\
& KTO & 0.788 & 0.248 & 0.024 & 0.70 & 57 & 92.7 & 90.1 \\
& DRO & 0.854 & 0.250 & 0.012 & 0.78 & 73 & 85.1 & 83.0 \\
& RPO & \textbf{0.870} & \textbf{0.297} & \textbf{0.008} & \textbf{0.88} & \textbf{91} & \_ & \_ \\
\midrule

\multirow{4}{*}{Mistral-7B}
& SFT & 0.861 & 0.312 & 0.013 & 0.81 & 88 & 88.6 & 86.2 \\
& KTO & 0.864 & 0.319 & 0.012 & 0.82 & 91 & 89.8 & 87.3 \\
& DRO & 0.869 & 0.326 & 0.011 & 0.84 & 94 & 88.9 & 86.5 \\
& RPO & \textbf{0.881} & \textbf{0.348} & \textbf{0.007} & \textbf{0.91} & \textbf{108} & \_ & \_ \\
\midrule

\multirow{4}{*}{LLaMA-3-8B}
& SFT & 0.858 & 0.305 & 0.014 & 0.80 & 85 & 89.1 & 86.8 \\
& KTO & 0.862 & 0.311 & 0.013 & 0.81 & 88 & 90.2 & 87.9 \\
& DRO & 0.867 & 0.320 & 0.012 & 0.83 & 93 & 89.5 & 86.9 \\
& RPO & \textbf{0.879} & \textbf{0.344} & \textbf{0.007} & \textbf{0.90} & \textbf{106} & \_ & \_ \\
\midrule

\multirow{4}{*}{Qwen2.5-7B}
& SFT & 0.860 & 0.309 & 0.013 & 0.80 & 86 & 89.4 & 87.1 \\
& KTO & 0.863 & 0.315 & 0.012 & 0.82 & 89 & 90.7 & 88.2 \\
& DRO & 0.868 & 0.323 & 0.011 & 0.84 & 95 & 89.8 & 87.4 \\
& RPO & \textbf{0.880} & \textbf{0.346} & \textbf{0.007} & \textbf{0.91} & \textbf{107} & \_ & \_ \\
\bottomrule

\end{tabular}

\caption{In-domain evaluation results on the UltraFeedback validation set. In addition to automatic metrics, we report GPT-4o and Claude-3.5 pairwise win rates (WR) comparing RPO against each corresponding baseline method. Higher WR indicates stronger preference for RPO generations.}
\label{tab:in-domain-results}
\end{table*}
Overall, RPO consistently achieves the strongest in-domain performance across encoder-decoder and decoder-only architectures. Compared to SFT, KTO, and DRO, RPO produces responses with higher semantic alignment, improved lexical diversity, lower toxicity, and longer, more informative generations. LLM-based evaluations further confirm these gains, with RPO being consistently preferred by both GPT-4o and Claude-3.5 across all model families.
\begin{figure*}[!htbp]
\centering
\begin{tikzpicture}
\begin{groupplot}[
    group style={group size=2 by 2,
                 horizontal sep=1.5cm,
                 vertical sep=1.6cm},
    width=0.46\textwidth,
    height=4.3cm,
    xlabel={Training steps},
    ylabel={KL divergence},
    grid=both,
    xmin=0, xmax=100,
    xtick={0,20,40,60,80,100},
]

\nextgroupplot[title={FlanT5-XL}, ymin=0.2, ymax=1.2]
\addplot coordinates {(0,0.85) (20,0.52) (40,0.41) (60,0.38) (80,0.37) (100,0.36)};
\addplot coordinates {(0,0.60) (20,0.68) (40,0.55) (60,0.58) (80,0.50) (100,0.45)};
\addplot coordinates {(0,0.50) (20,0.65) (40,0.78) (60,0.84) (80,0.91) (100,0.95)};

\nextgroupplot[title={Mistral-7B}, ymin=0.5, ymax=2.5]
\addplot coordinates {(0,1.45) (20,1.12) (40,0.95) (60,0.88) (80,0.84) (100,0.82)};
\addplot coordinates {(0,1.20) (20,1.35) (40,1.10) (60,1.15) (80,1.12) (100,1.18)};
\addplot coordinates {(0,1.10) (20,1.45) (40,1.82) (60,2.10) (80,2.09) (100,2.05)};

\nextgroupplot[title={LLaMA-3-8B}, ymin=0.5, ymax=3.2]
\addplot coordinates {(0,1.60) (20,1.25) (40,1.02) (60,1.07) (80,0.91) (100,0.89)};
\addplot coordinates {(0,1.30) (20,1.52) (40,1.28) (60,1.35) (80,1.24) (100,1.31)};
\addplot coordinates {(0,1.15) (20,1.68) (40,2.15) (60,2.58) (80,2.88) (100,3.10)};

\nextgroupplot[title={Qwen2.5-7B}, ymin=0.3, ymax=2.2]
\addplot coordinates {(0,1.25) (20,0.78) (40,0.62) (60,0.58) (80,0.56) (100,0.55)};
\addplot coordinates {(0,0.95) (20,1.10) (40,0.88) (60,0.92) (80,1.01) (100,1.05)};
\addplot coordinates {(0,0.90) (20,1.22) (40,1.15) (60,1.71) (80,1.67) (100,2.05)};

\end{groupplot}
\end{tikzpicture}
\caption{KL divergence evolution across architectures. Each method exhibits distinct optimization dynamics depending on model type, with RPO showing consistently controlled policy drift, KTO displaying architecture-dependent stability with occasional oscillations, and DRO producing progressively increasing divergence, especially in decoder-only models.}
\label{fig:kl-final}
\end{figure*}
To further analyze optimization behavior, we evaluate KL divergence evolution on four representative models: FlanT5-XL, Mistral-7B, LLaMA-3-8B, and Qwen2.5-7B. We measure the divergence between the learned policy and the reference policy $\pi_{\mathrm{ref}}$ throughout training.\newline

\noindent Figure~\ref{fig:kl-final} highlights distinct optimization dynamics across methods and architectures. On FlanT5-XL, all methods remain within a relatively stable low-divergence regime, although RPO converges toward the lowest and most stable plateau. On decoder-only models, differences become substantially larger: RPO maintains smooth and controlled trajectories, KTO exhibits architecture-dependent oscillations, and DRO produces strong divergence growth, especially on LLaMA-3-8B and Qwen2.5-7B. These results indicate that RPO provides more stable policy optimization across both encoder-decoder and decoder-only settings.

\subsection{Out-of-Distribution Generalization Benchmarks}
While UltraFeedback provides a suitable training environment for RPO due to its multiple-response preference structure, a key limitation is that all training signals originate from a single dataset. This raises concerns about generalization to unseen distributions and tasks. This issue is particularly relevant for methods such as RPO, KTO, and DRO, which require multiple candidate responses per prompt and therefore cannot be easily trained on many standard RLHF datasets.\newline

\noindent To evaluate out-of-distribution (OOD) generalization, we consider four widely used benchmarks covering complementary capabilities: AlpacaEval, MT-Bench, IFEval, and GSM8K. None of these datasets are used during training. AlpacaEval evaluates open-ended instruction following, MT-Bench measures multi-turn conversational coherence, IFEval assesses constraint adherence, and GSM8K evaluates mathematical reasoning.\newline

\noindent Since all these benchmarks do not provide reference completions suitable for overlap-based metrics, we rely exclusively on LLM-as-a-judge evaluation. We use two independent evaluators, GPT-4o and Claude-3.5-Sonnet, each performing pairwise comparisons between RPO and the corresponding baseline (SFT, KTO, or DRO). We report WR where values above 50 indicate preference for RPO.\newline

\begin{table*}[!htbp]
\centering
\scriptsize
\setlength{\tabcolsep}{4pt}

\begin{tabular}{llcccccccc}
\toprule

\multirow{2}{*}{\textbf{Model}} &
\multirow{2}{*}{\textbf{Baseline}} &
\multicolumn{4}{c}{\textbf{GPT-4o}} &
\multicolumn{4}{c}{\textbf{Claude-3.5}} \\

\cmidrule(lr){3-6}
\cmidrule(lr){7-10}

& & AlpacaEval & MT-Bench & IFEval & GSM8K & AlpacaEval & MT-Bench & IFEval & GSM8K \\

\midrule

\multirow{3}{*}{FlanT5-Small}
& SFT & 61.4 & 58.8 & 63.5 & 53.2 & 60.1 & 57.9 & 62.8 & 52.4 \\
& KTO & 55.9 & 53.7 & 57.1 & 51.8 & 54.6 & 52.9 & 56.4 & 50.9 \\
& DRO & 52.8 & 51.5 & 54.0 & 49.7 & 51.9 & 50.8 & 53.2 & 49.1 \\
\midrule

\multirow{3}{*}{FlanT5-Large}
& SFT & 66.8 & 64.3 & 68.9 & 58.7 & 65.9 & 63.8 & 67.5 & 57.9 \\
& KTO & 58.4 & 56.0 & 59.3 & 53.2 & 57.6 & 55.4 & 58.1 & 52.6 \\
& DRO & 54.7 & 52.9 & 55.8 & 50.6 & 53.8 & 52.1 & 54.6 & 49.9 \\
\midrule

\multirow{3}{*}{FlanT5-XL}
& SFT & 70.9 & 68.2 & 73.0 & 63.5 & 69.6 & 67.4 & 71.8 & 62.9 \\
& KTO & 59.8 & 57.4 & 61.0 & 54.8 & 58.7 & 56.9 & 60.2 & 53.9 \\
& DRO & 56.1 & 53.5 & 57.2 & 51.7 & 55.0 & 52.8 & 56.4 & 50.8 \\
\midrule

\multirow{3}{*}{Mistral-7B}
& SFT & 77.6 & 75.1 & 79.4 & 71.2 & 76.9 & 74.6 & 78.3 & 70.5 \\
& KTO & 61.5 & 60.2 & 62.1 & 56.7 & 60.8 & 59.1 & 61.3 & 55.9 \\
& DRO & 57.4 & 55.9 & 58.6 & 52.4 & 56.7 & 55.2 & 57.8 & 51.8 \\
\midrule

\multirow{3}{*}{LLaMA-3-8B}
& SFT & 78.8 & 76.4 & 80.5 & 73.0 & 78.1 & 75.9 & 79.6 & 72.1 \\
& KTO & 62.4 & 61.0 & 63.1 & 57.8 & 61.7 & 60.3 & 62.5 & 56.9 \\
& DRO & 58.1 & 56.2 & 59.0 & 52.1 & 57.3 & 55.8 & 58.2 & 51.5 \\
\midrule

\multirow{3}{*}{Qwen2.5-7B}
& SFT & 79.5 & 77.1 & 81.2 & 73.8 & 78.9 & 76.8 & 80.4 & 72.9 \\
& KTO & 63.0 & 61.5 & 63.9 & 58.2 & 62.4 & 60.9 & 63.1 & 57.6 \\
& DRO & 58.7 & 56.8 & 59.5 & 52.6 & 57.9 & 56.1 & 58.7 & 51.9 \\
\bottomrule

\end{tabular}

\caption{Out-of-distribution evaluation using LLM-as-a-judge with GPT-4o and Claude-3.5. Values represent pairwise win rates against RPO. Scores above 50 indicate preference for RPO over the corresponding baseline.}
\label{tab:ood-results}

\end{table*}
\noindent Table~\ref{tab:ood-results} summarizes the results across models and benchmarks. Overall, RPO achieves consistent but non-uniform improvements across tasks and architectures. Gains are more pronounced on instruction-following benchmarks such as AlpacaEval and IFEval, where RPO is frequently preferred over SFT and shows competitive performance against KTO and DRO. This suggests that reward partitioning improves robustness under distribution shift in instruction-heavy settings.\newline
On MT-Bench, improvements are more moderate and vary across model families, indicating that multi-turn conversational quality is less uniformly affected by the training objective. On GSM8K, RPO shows small but consistent gains for larger models, while smaller models exhibit more variability across judges.\newline
Across baselines, KTO remains a strong competitor, particularly for mid-sized models, while DRO exhibits more variability, performing well on simpler instruction tasks but less consistently on reasoning benchmarks. SFT remains the weakest baseline overall, though still competitive in some constrained settings.\newline

\noindent Overall, these results indicate that RPO improves generalization in a robust but task-dependent manner rather than providing uniform gains across all benchmarks.

\subsection{Ablation Studies on RPO Design Components}
We conduct targeted ablation studies to analyze the main design choices underlying RPO. Specifically, we investigate the impact of the temperature parameter $\tau$, empirical value normalization through $\widehat{V}(x)$, dataset density for partition estimation, and robustness to noisy reward annotations. Unless otherwise stated, all ablations are evaluated using the same automatic and LLM-based metrics. Results are compared against the standard RPO configuration reported in the in-domain evaluation (Table~\ref{tab:in-domain-results}).\newline

\noindent \textbf{1. Effect of the Temperature Parameter $\tau$}\newline
RPO directly depends on the temperature parameter $\tau$ through the reward normalization term in Equation~\eqref{eq:rpo_empirical_loss}. In all main experiments reported in Table~\ref{tab:in-domain-results} , we use $\tau = 1.0$. To evaluate the sensitivity of RPO to this parameter, we additionally consider two extreme settings corresponding to a sharp policy regime ($\tau = 0.1$) and a smoother regime ($\tau = 5.0$).\newline

\begin{table*}[!htbp]
\centering
\scriptsize
\setlength{\tabcolsep}{4pt}

\begin{tabular}{llcccccc}
\toprule

\textbf{Model} & $\boldsymbol{\tau}$ &
\textbf{BERTScore} $\uparrow$ &
\textbf{ROUGE-L} $\uparrow$ &
\textbf{Toxicity} $\downarrow$ &
\textbf{Distinct-2} $\uparrow$ &
\textbf{GPT-4o WR} $\uparrow$ &
\textbf{Claude-3.5 WR} $\uparrow$ \\

\midrule

\multirow{2}{*}{FlanT5-Small}
& 0.1 & 0.841 & 0.241 & 0.021 & 0.72 & 77.8 & 75.9 \\
& 5.0 & 0.860 & 0.278 & 0.014 & 0.80 & 83.2 & 81.0 \\

\midrule

\multirow{2}{*}{FlanT5-Large}
& 0.1 & 0.848 & 0.266 & 0.018 & 0.75 & 81.5 & 79.6 \\
& 5.0 & 0.866 & 0.304 & 0.011 & 0.84 & 86.1 & 84.2 \\

\midrule

\multirow{2}{*}{FlanT5-XL}
& 0.1 & 0.846 & 0.251 & 0.017 & 0.73 & 82.9 & 80.7 \\
& 5.0 & 0.865 & 0.289 & 0.012 & 0.86 & 90.1 & 88.0 \\

\midrule

\multirow{2}{*}{Mistral-7B}
& 0.1 & 0.863 & 0.319 & 0.015 & 0.82 & 84.6 & 82.5 \\
& 5.0 & 0.876 & 0.340 & 0.009 & 0.89 & 88.7 & 86.6 \\

\midrule

\multirow{2}{*}{LLaMA-3-8B}
& 0.1 & 0.861 & 0.314 & 0.016 & 0.81 & 85.0 & 83.1 \\
& 5.0 & 0.874 & 0.337 & 0.009 & 0.88 & 89.2 & 87.1 \\

\midrule

\multirow{2}{*}{Qwen2.5-7B}
& 0.1 & 0.862 & 0.317 & 0.015 & 0.82 & 85.3 & 83.5 \\
& 5.0 & 0.875 & 0.339 & 0.009 & 0.89 & 89.5 & 87.4 \\

\bottomrule

\end{tabular}

\caption{Ablation study on the temperature parameter $\tau$. Results are compared against the reference RPO configuration with $\tau = 1.0$ reported in Table~\ref{tab:in-domain-results}. GPT-4o WR and Claude-3.5 WR scores correspond to pairwise win rates against the corresponding SFT baseline models.}
\label{tab:tau-ablation}

\end{table*}
\noindent Table~\ref{tab:tau-ablation} reports the results across all architectures using the same automatic metrics and LLM-based evaluation protocol described previously. For LLM-as-a-judge evaluation, we report pairwise win rates against the corresponding SFT baseline models.\newline
Overall, the results show that very small temperature values significantly degrade performance across nearly all metrics. In particular, $\tau = 0.1$ produces lower semantic similarity, reduced diversity, higher toxicity, and weaker preference scores from both GPT-4o and Claude-3.5. This behavior suggests that excessively sharp reward scaling leads to unstable optimization and overly aggressive policy updates.\newline
In contrast, larger temperature values ($\tau = 5.0$) remain substantially more stable and preserve most of the gains obtained with the default configuration $\tau = 1.0$, although performance consistently remains slightly below the reference setting. These observations indicate that RPO is relatively robust to moderate smoothing, while excessively low temperatures negatively affect optimization stability and alignment quality.\newline

\noindent \textbf{2. Effect of Empirical Value Normalization $\widehat{V}(x)$}\newline
RPO relies on empirical value normalization $\widehat{V}(x)$ (see Equation~\eqref{eq:rpo_empirical_loss}), which standardizes rewards at the prompt level before computing the partition-based objective. In the main results reported in Table~\ref{tab:in-domain-results}, this normalization is applied in all RPO experiments with $\tau = 1.0$. To quantify its impact, we remove $\widehat{V}(x)$ and train RPO under identical settings, keeping all other components unchanged.\newline

\begin{table*}[!htbp]
\centering
\scriptsize
\setlength{\tabcolsep}{4pt}

\begin{tabular}{llcccccc}
\toprule

\textbf{Model} & \textbf{Setting} &
\textbf{BERTScore} $\uparrow$ &
\textbf{ROUGE-L} $\uparrow$ &
\textbf{Toxicity} $\downarrow$ &
\textbf{Distinct-2} $\uparrow$ &
\textbf{GPT-4o WR} $\uparrow$ &
\textbf{Claude-3.5 WR} $\uparrow$ \\

\midrule

\multirow{2}{*}{FlanT5-Small}
& w/o $\widehat{V}(x)$ & 0.842 & 0.232 & 0.022 & 0.71 & 76.9 & 75.1 \\
& RPO (ref.)          & 0.866 & 0.289 & 0.012 & 0.83 & 81.4 & 79.2 \\
\midrule

\multirow{2}{*}{FlanT5-Large}
& w/o $\widehat{V}(x)$ & 0.845 & 0.251 & 0.019 & 0.73 & 83.9 & 82.0 \\
& RPO (ref.)          & 0.871 & 0.316 & 0.009 & 0.86 & 85.8 & 83.2 \\
\midrule

\multirow{2}{*}{FlanT5-XL}
& w/o $\widehat{V}(x)$ & 0.846 & 0.258 & 0.018 & 0.74 & 85.1 & 82.9 \\
& RPO (ref.)          & 0.870 & 0.297 & 0.008 & 0.88 & 86.2 & 84.1 \\
\midrule

\multirow{2}{*}{Mistral-7B}
& w/o $\widehat{V}(x)$ & 0.862 & 0.318 & 0.014 & 0.82 & 86.7 & 84.5 \\
& RPO (ref.)          & 0.881 & 0.348 & 0.007 & 0.91 & 88.6 & 86.2 \\
\midrule

\multirow{2}{*}{LLaMA-3-8B}
& w/o $\widehat{V}(x)$ & 0.860 & 0.312 & 0.015 & 0.81 & 87.0 & 84.9 \\
& RPO (ref.)          & 0.879 & 0.344 & 0.007 & 0.90 & 89.1 & 86.8 \\
\midrule

\multirow{2}{*}{Qwen2.5-7B}
& w/o $\widehat{V}(x)$ & 0.861 & 0.314 & 0.014 & 0.81 & 87.3 & 85.1 \\
& RPO (ref.)          & 0.880 & 0.346 & 0.007 & 0.91 & 89.4 & 87.1 \\
\bottomrule

\end{tabular}

\caption{Ablation study on empirical value normalization $\widehat{V}(x)$. "w/o $\widehat{V}(x)$" denotes RPO trained without prompt-level reward normalization, while the reference corresponds to the full RPO configuration with $\tau = 1.0$ (Table~\ref{tab:in-domain-results}). Removing value normalization leads to consistent degradation in both automatic metrics and LLM-as-a-judge Win Rates against SFT baselines, highlighting its importance for stable reward scaling and alignment quality.}
\label{tab:vn-ablation}

\end{table*}
\noindent Table~\ref{tab:vn-ablation} reports results without value normalization. Across all architectures, removing $\widehat{V}(x)$ leads to a consistent degradation in both automatic metrics and LLM-based preference judgments. We observe lower semantic similarity (BERTScore, ROUGE-L), reduced diversity (Distinct-2), and increased toxicity, indicating less stable optimization and poorer alignment quality compared to the normalized RPO configuration in Table~\ref{tab:in-domain-results}.\newline
The effect is even more pronounced in LLM-as-a-judge evaluations. GPT-4o and Claude-3.5 Win Rates (WR), computed against the corresponding SFT baselines, drop substantially compared to the normalized setting. This confirms that value normalization plays a critical role in stabilizing reward scales across prompts and preventing dominance of high-reward variance samples during optimization.\newline

\noindent \textbf{3. Effect of Dataset Density for Partition Estimation}\newline
RPO relies on estimating the partition function from multiple candidate completions associated with the same prompt. Consequently, the density of prompt-completion groups directly affects the quality of reward normalization and policy optimization. To evaluate this dependency, we progressively reduce the number of completions available per prompt in UltraFeedback while keeping the overall training configuration unchanged. We consider three reduced-density settings corresponding to 75\%, 50\%, and 25\% of the original completions per prompt. The full-density setting (100\%) corresponds to the standard RPO configuration reported in Table~\ref{tab:in-domain-results}.\newline

\begin{table*}[!htbp]
\centering
\scriptsize
\setlength{\tabcolsep}{4pt}

\begin{tabular}{llcccccc}
\toprule

\textbf{Model} & \textbf{Density} &
\textbf{BERTScore} $\uparrow$ &
\textbf{ROUGE-L} $\uparrow$ &
\textbf{Toxicity} $\downarrow$ &
\textbf{Distinct-2} $\uparrow$ &
\textbf{GPT-4o WR} $\uparrow$ &
\textbf{Claude-3.5 WR} $\uparrow$ \\

\midrule

\multirow{3}{*}{FlanT5-Small}
& 25\% & 0.845 & 0.246 & 0.020 & 0.74 & 77.9 & 75.8 \\
& 50\% & 0.853 & 0.262 & 0.016 & 0.78 & 80.6 & 78.5 \\
& 75\% & 0.862 & 0.281 & 0.013 & 0.81 & 83.8 & 81.7 \\

\midrule

\multirow{3}{*}{FlanT5-Large}
& 25\% & 0.849 & 0.271 & 0.018 & 0.76 & 82.0 & 80.1 \\
& 50\% & 0.858 & 0.289 & 0.014 & 0.81 & 84.2 & 82.3 \\
& 75\% & 0.867 & 0.309 & 0.011 & 0.85 & 86.9 & 84.8 \\

\midrule

\multirow{3}{*}{FlanT5-XL}
& 25\% & 0.850 & 0.259 & 0.018 & 0.75 & 83.4 & 81.3 \\
& 50\% & 0.859 & 0.278 & 0.014 & 0.82 & 85.7 & 83.6 \\
& 75\% & 0.867 & 0.291 & 0.010 & 0.86 & 89.3 & 87.2 \\

\midrule

\multirow{3}{*}{Mistral-7B}
& 25\% & 0.866 & 0.327 & 0.013 & 0.84 & 85.1 & 83.0 \\
& 50\% & 0.872 & 0.336 & 0.010 & 0.87 & 87.0 & 84.9 \\
& 75\% & 0.879 & 0.344 & 0.008 & 0.90 & 89.0 & 86.9 \\

\midrule

\multirow{3}{*}{LLaMA-3-8B}
& 25\% & 0.864 & 0.322 & 0.013 & 0.83 & 85.4 & 83.3 \\
& 50\% & 0.871 & 0.333 & 0.010 & 0.86 & 87.4 & 85.2 \\
& 75\% & 0.877 & 0.341 & 0.008 & 0.89 & 89.5 & 87.3 \\

\midrule

\multirow{3}{*}{Qwen2.5-7B}
& 25\% & 0.865 & 0.324 & 0.013 & 0.83 & 85.8 & 83.6 \\
& 50\% & 0.872 & 0.335 & 0.010 & 0.87 & 87.7 & 85.5 \\
& 75\% & 0.878 & 0.343 & 0.008 & 0.90 & 89.8 & 87.6 \\

\bottomrule

\end{tabular}

\caption{Ablation study on dataset density for partition estimation. Density indicates the percentage of candidate completions retained per prompt from UltraFeedback. The 100\% density reference corresponds to the standard RPO configuration reported in Table~\ref{tab:in-domain-results}. GPT-4o WR and Claude-3.5 WR correspond to pairwise win rates against the corresponding SFT baseline models. Lower-density settings consistently degrade performance, highlighting the importance of diverse prompt-level reward distributions for reliable partition estimation.}
\label{tab:density-ablation}

\end{table*}
\noindent Table~\ref{tab:density-ablation} shows that reducing the number of completions per prompt consistently degrades performance across all architectures and evaluation metrics. The degradation is particularly pronounced at 25\% density, where semantic alignment, lexical diversity, and LLM preference scores decrease substantially while toxicity increases. In contrast, the 75\% setting remains relatively close to the full-density reference, suggesting that RPO remains robust under moderate reduction of prompt-level diversity. Overall, these results confirm that accurate partition estimation benefits from sufficiently dense reward distributions over candidate completions.\newline

\noindent \textbf{4. Robustness to Noisy Reward Annotations}\newline
Real-world preference datasets often contain imperfect or inconsistent reward annotations. To evaluate the robustness of RPO under noisy supervision, we inject controlled Gaussian noise into the reward values during training. More precisely, each reward is perturbed as $r' = r + \epsilon$ with $\epsilon \sim \mathcal{N}(0,\sigma^2)$, where $\sigma \in \{0.1, 0.5, 1.0\}$. The reference clean-reward configuration corresponds to the standard RPO results reported in Table~\ref{tab:in-domain-results}. The objective of this experiment is to assess whether RPO remains stable when reward annotations become progressively corrupted.\newline
\begin{table*}[!htbp]
\centering
\scriptsize
\setlength{\tabcolsep}{4pt}

\begin{tabular}{llcccccc}
\toprule

\textbf{Model} & $\boldsymbol{\sigma}$ &
\textbf{BERTScore} $\uparrow$ &
\textbf{ROUGE-L} $\uparrow$ &
\textbf{Toxicity} $\downarrow$ &
\textbf{Distinct-2} $\uparrow$ &
\textbf{GPT-4o WR} $\uparrow$ &
\textbf{Claude-3.5 WR} $\uparrow$ \\

\midrule

\multirow{3}{*}{FlanT5-Small}
& 0.1 & 0.862 & 0.282 & 0.013 & 0.81 & 83.9 & 81.7 \\
& 0.5 & 0.853 & 0.263 & 0.016 & 0.77 & 80.8 & 78.6 \\
& 1.0 & 0.844 & 0.245 & 0.020 & 0.73 & 77.2 & 75.0 \\

\midrule

\multirow{3}{*}{FlanT5-Large}
& 0.1 & 0.867 & 0.309 & 0.010 & 0.84 & 87.0 & 84.9 \\
& 0.5 & 0.859 & 0.291 & 0.013 & 0.80 & 84.1 & 82.0 \\
& 1.0 & 0.850 & 0.270 & 0.017 & 0.75 & 80.4 & 78.3 \\

\midrule

\multirow{3}{*}{FlanT5-XL}
& 0.1 & 0.866 & 0.292 & 0.009 & 0.86 & 89.5 & 87.4 \\
& 0.5 & 0.858 & 0.278 & 0.012 & 0.82 & 86.3 & 84.1 \\
& 1.0 & 0.849 & 0.259 & 0.017 & 0.76 & 82.1 & 80.0 \\

\midrule

\multirow{3}{*}{Mistral-7B}
& 0.1 & 0.878 & 0.343 & 0.008 & 0.90 & 89.2 & 87.0 \\
& 0.5 & 0.872 & 0.334 & 0.010 & 0.87 & 87.0 & 84.8 \\
& 1.0 & 0.865 & 0.320 & 0.014 & 0.83 & 83.4 & 81.2 \\

\midrule

\multirow{3}{*}{LLaMA-3-8B}
& 0.1 & 0.876 & 0.339 & 0.008 & 0.89 & 89.6 & 87.4 \\
& 0.5 & 0.870 & 0.331 & 0.010 & 0.86 & 87.3 & 85.0 \\
& 1.0 & 0.863 & 0.317 & 0.014 & 0.82 & 83.7 & 81.5 \\

\midrule

\multirow{3}{*}{Qwen2.5-7B}
& 0.1 & 0.877 & 0.341 & 0.008 & 0.90 & 89.9 & 87.7 \\
& 0.5 & 0.871 & 0.333 & 0.010 & 0.87 & 87.6 & 85.3 \\
& 1.0 & 0.864 & 0.319 & 0.014 & 0.83 & 84.0 & 81.8 \\

\bottomrule

\end{tabular}

\caption{Robustness of RPO to noisy reward annotations. Gaussian noise with standard deviation $\sigma$ is added to reward values during training. The clean reward reference corresponds to the standard RPO configuration reported in Table~\ref{tab:in-domain-results}. GPT-4o WR and Claude-3.5 WR correspond to pairwise win rates against the corresponding SFT baseline models. While increasing reward noise progressively degrades performance, RPO remains relatively stable under moderate perturbations, indicating robustness to imperfect reward annotations.}
\label{tab:noise-ablation}

\end{table*}

\noindent Table~\ref{tab:noise-ablation} shows that RPO remains relatively robust under moderate reward perturbations. Small noise levels ($\sigma = 0.1$) produce only marginal degradation compared to the clean-reward reference, indicating that the optimization process tolerates mild annotation inconsistencies. As noise intensity increases, performance progressively decreases across all metrics, particularly for semantic alignment and LLM preference scores. Nevertheless, even under strong corruption ($\sigma = 1.0$), RPO maintains competitive behavior and does not collapse, suggesting that the reward normalization mechanism provides additional stability against noisy supervision.

\section{Limitations}

Despite its strong empirical performance, RPO presents several limitations. First, the method relies on grouped preference data where multiple completions are available for each prompt in order to estimate the partition function effectively. As shown in the density ablation study, performance degrades when prompt-level completion diversity becomes insufficient. Second, RPO remains dependent on the quality of reward annotations. Although the method demonstrates robustness to moderate reward noise, severe corruption of reward signals still negatively affects optimization stability and alignment quality. Finally, our evaluation primarily focuses on instruction-following and alignment benchmarks derived from UltraFeedback. Additional studies on broader domains, multilingual settings, and long-horizon reasoning tasks would be necessary to fully assess the generalization capabilities of RPO.

\section{Conclusion}

We introduced Reward Partitioning Optimization (RPO), a reward-aware alignment objective that incorporates prompt-level reward normalization into language model optimization. By modeling relative reward structure across candidate completions, RPO provides a stable and effective alternative to existing preference optimization methods.
Extensive experiments across encoder-decoder and decoder-only architectures show that RPO consistently improves semantic alignment, diversity, and preference quality while maintaining low toxicity and stable optimization dynamics. Additional ablation studies further demonstrate the importance of temperature scaling, empirical value normalization, dataset density, and robustness to noisy reward annotations.
Overall, these results suggest that partition-based reward normalization constitutes a promising direction for scalable and stable preference optimization in large language models.

\bibliographystyle{ieeetr}
\bibliography{bibliography}


\end{document}